\documentclass{article}
\usepackage{ijcai21}

\usepackage{times}

\usepackage{soul}
\usepackage{url}
\usepackage[hidelinks]{hyperref}
\usepackage[utf8]{inputenc}
\usepackage[small]{caption}
\usepackage{graphicx}
\usepackage{amsmath}
\usepackage{booktabs}
\usepackage{xcolor}

\pdfoutput=1 

\title{International Workshop on Continual Semi-Supervised Learning:\\ Introduction, Benchmarks and Baselines}

\author
{
Ajmal Shahbaz$^1$\footnote{Contact Author}\and
Salman Khan$^1$ \and
Mohammad Asiful Hossain$^2$\and
Vincenzo Lomonaco$^{3}$\and
Kevin Cannons$^2$
\and
Zhan Xu$^2$
\And
Fabio Cuzzolin$^1$\\
\affiliations
$^1$Oxford Brookes University\\
$^2$Huawei Technologies\\
$^3$University of Pisa
\emails
\{ashahbaz, salmankhan, fabio.cuzzolin\}@brookes.ac.uk,
\{mohammad.asiful.hossain, kevin.cannons, zhan.xu\}@huawei.com,
vincenzo.lomonaco@unipi.it
}

\begin{document}

\maketitle

\begin{abstract}

The aim of this paper is to formalise a new continual semi-supervised learning (CSSL) paradigm, proposed to the attention of the machine learning community via the IJCAI 2021 International Workshop on Continual Semi-Supervised Learning (CSSL@IJCAI)\footnote{\url{https://sites.google.com/view/sscl-workshop-ijcai-2021/}}, 
with the aim of raising the field's awareness about this problem and mobilising its effort in this direction. 
After a formal definition of continual semi-supervised learning and the appropriate training and testing protocols,
the paper introduces two new benchmarks specifically designed to assess CSSL on two important computer vision tasks: activity recognition and crowd counting. We describe the Continual Activity Recognition (CAR) and Continual Crowd Counting (CCC) challenges built upon those benchmarks, the baseline models proposed for the challenges, and describe a simple CSSL baseline which consists in applying batch self-training in temporal sessions, for a limited number of rounds. The results show that learning from unlabelled data streams is extremely challenging, and stimulate the search for methods that can encode the dynamics of the data stream. 
\end{abstract}

\section{Introduction}

Whereas the continual learning problem has been recently the object of much attention in the machine learning community, the latter has mainly studied from the point of view of class-incremental learning \cite{van2019three}, in particular with a focus on preventing the model updated in the light of new data from `catastrophic forgetting' its initial, useful knowledge and abilities.
A typical example is that of an object detector which needs to be extended to include classes not originally in its list (e.g., `donkey') while retaining its ability to correctly detect, say, a `horse'.
The hidden assumption there is that we are quite satisfied with the model we have, and we merely wish to extend its capabilities to new settings and classes \cite{prabhu2020gdumb}.

This way of posing the continual learning problem, however, is in rather stark contrast with widespread real-world situations in which an initial model is trained using limited data, only for it to then be deployed without any additional supervision.
Think of a detector used for traffic monitoring on a busy street. Even after having been trained extensively on the many available public datasets, experience shows that its performance in its target setting will likely be less than optimal. In this scenario, the objective is for the model to be incrementally updated using the new (unlabelled) data, in order to adapt to its new target domain.
\\
In such settings the goal is quite the opposite to that of the classical scenario described above: 
the initial model is usually rather poor and not a good match for the target domain.
To complicate things further, the target domain itself may change over time, both periodically (think of night/day cycles) and in asynchronous, discrete steps (e.g., when a new bank opens within the camera's field of view, or a new building is erected with a new entrance) \cite{bitarafan2016incremental}.

In this paper we formalise this problem as one of \emph{continual semi-supervised learning}, in which an initial training batch of labelled data points is available and can be used to train an initial model, but then the model is incrementally updated exploiting the information provided by a time series of unlabelled data points, each of which is generated by a data generating process (modelled by a probability distribution) which \emph{varies with time}.
We do not assume any artificial subdivision into `tasks', but allow the data generating distribution to be an arbitrary function of time.

While batch semi-supervised learning (SSL) has seen renewed interest in recent times, thanks to relevant work in the rising field of unsupervised domain adaptation \cite{Tzeng2017adversarial} and various approaches based on classical SSL self-training \cite{rosenberg2005semi,chen2019progressive}, 
continual SSL is still a rather unexplored field.
The reason is that, while in the supervised case it is clear what information the streaming data points carry, in the semi-supervised case it is far from obvious what relevant information carried by the streaming instance should drive model update.

\subsection{Contributions}

With this paper we wish to contribute:
\begin{enumerate}
\item
A formal definition of the continual semi-supervised learning problem, and the associated training and testing protocols, as a basis for future work in this area.
\item
The first two benchmark datasets for the validation of semi-supervised continual learning approaches, one for classification (continual activity recognition, CAR) and one for regression (continual crowd counting, CCC), which we propose as the foundations of the first challenges in this domain.
\item
Results produced by a simple strategy in which self-training is applied in sessions to competitive baseline models on both benchmarks, as the foundations for the above challenges.
\end{enumerate}
Our results confirm that continual semi-supervised learning is a very challenging task, and that strategies that fail to model the dynamics of the unlabelled data stream have limited performance potential.

\subsection{Paper outline}

The structure of the paper is as follows. First we discuss some relevant related work in continual supervised learning and semi-supervised learning (Section \ref{sec:related}). We then introduce the principle and framework of our continual semi-supervised learning approach (Section \ref{sec:continual-ssl}), in particular the relevant training and testing protocols.
In Section \ref{sec:benchmarks} we introduce our new benchmark datasets, while in Section \ref{sec:challenges} we illustrate the IJCAI 2021 challenges based upon those datasets and the activity detection and crowd counting approaches we adopted as baselines, together with our simple self-training strategy for tackling the challenge. In Section \ref{sec:experiments} we illustrate our baseline results for the two challenges. Section \ref{sec:conclusions} concludes.

\section{Related work} \label{sec:related}

\subsection{Continual supervised learning} \label{sec:related-supervised}

\emph{Continual learning} refers to the ability of a system to keep learning from new data throughout its lifetime, even beyond an initial training stage, allowing the system to adapt to new, ever evolving situations (‘domain adaptation’). Continual learning can be supervised, when new annotation is provided even after the system is deployed, or unsupervised/semi-supervised if, after training on an initial batch of annotated data-points (e.g., videos with an attached ‘event’ label, or images with superimposed manually drawn skeletons of the humans in the scene), the data subsequently captured cannot be annotated. In supervised continual learning, the focus of the community has been so far on the issue of avoiding ‘catastrophic forgetting’, i.e., models that forget their previous abilities and they keep being updated in the light of new data. 

Current continual learning methods can be categorised into three major families based on how the information of previous data are stored and used. \emph{Prior-focused} methods \cite{Farquhar2018TowardsRE} use a penalty term to regularise the parameters rather than a hard constraint. \emph{Parameter isolation} methods \cite{Mallya2018PackNetAM} dedicate different parameters for different tasks to prevent interference. Finally, \emph{replay-based} approaches store the examples’ information in a ‘replay buffer’ or a generative model, which is used for rehearsal/retraining or to provide constraints on the current learning step \cite{Aljundi2019online}.


\subsection{Continual learning as constrained optimisation}

In particular, in a recent paper \cite{Aljundi2019online}, continual supervised learning is formulated in terms of the following optimization problem, for any time instant $t$:
$\theta_t = \arg\min_\theta l (f(x_t|\theta), y_t),$
subject to the constraints that 
$l ( f( x_\tau | \theta) , y_\tau ) \leq l ( f( x_\tau | \theta_{t-1} ) , y_\tau ), \quad \forall \tau = 1, \ldots, t-1$,
where $y_\tau$ is the true label of instance $x_\tau$, and $\theta_t$ is the version of the model at time $t$. The loss function $l(y',y)$ assesses how well the predicted label $y'$ matches the actual label $y$. 
\\
The aim of supervised continual learning as formulated above 
is to incrementally learn a model able to best adapt to each new data-point, 
while ensuring that the loss of the new model on the past observed data-points is no greater than the loss of the previous version of the model there. 
Note that not all cost functions in model learning correspond to the average of some loss over the training examples. A (hidden) assumption in \cite{Aljundi2019online} is, therefore, that learning happens by minimising this average (`empirical loss').

The problem is designed to {avoid catastrophic forgetting} at instance (rather than class) level, as new models are required to have a non-increasing loss on the instances seen up to then.


\subsection{Unsupervised domain adaptation} \label{sec:related-uda}

Batch semi-supervised learning (SSL) has seen renewed interest in recent times, thanks to relevant work in the rising field of \emph{unsupervised domain adaptation}, where the dominant trend is based on adversarial approaches to reduce the discrepancy between source and target domain features \cite{Chen2018DomainAF,Tzeng2017AdversarialDD}. Many deep self-training approaches have been proposed, inspired by the classical SSL self-training algorithm, which alternate between generating pseudo-labels for the unlabelled data-points and re-training a deep classifier (including the intermediate layers producing the feature embeddings) \cite{Chen2019ProgressiveFA}. Despite all this, {continual semi-supervised learning} remains a rather unexplored field and  \cite{langeet}.


\section{Continual semi-supervised learning} \label{sec:continual-ssl}

We begin by formalising the continual semi-supervised learning problem, and by defining the appropriate training and testing protocols for this new learning setting.

\subsection{Learning} \label{sec:problem-learning}

The \emph{continual semi-supervised learning} problem can be formulated as follows.

Given an initial batch of supervised training examples, 
\[
\mathcal{T}_0 = \{ (x^j,y^j), j = 1, \ldots, N \}, 
\]
and a stream of unsupervised examples, 
\[
\mathcal{T}_t = \{ x_1, \ldots, x_t, \ldots \}, 
\]
we wish to learn a model $f(x|\theta)$ mapping instances $x \in \mathcal{X}$ to 
target values $y \in \mathcal{Y}$, 
depending on a vector $\theta \in \Theta$ of parameters. 

When $\mathcal{Y}$ is discrete (i.e., a list of labels) the problem is one of continual semi-supervised \emph{classification}; when $\mathcal{Y}$ is continuous we talk about continual semi-supervised \emph{regression}.

In both cases we wish the model to be updated incrementally as the string of unsupervised samples come in: namely, based on the current instance $x_t$ at time $t$, model $\theta_{t-1}$ is mapped to a new model $\theta_{t}$.

\subsection{Testing} \label{sec:problem-testing}

How should such a series of models be evaluated?
In continual supervised learning (see e.g. \cite{Aljundi2019online}), a series of ground-truth target values is available and can be exploited to update the model. 
To avoid overfitting, the series of models outputted by a continual supervised learner, $f(x|\theta_1), \ldots, f(x|\theta_t), \ldots$ cannot be tested on the same data it was trained upon. 
Thus, basically all continual (supervised) learning papers set aside from the start a set of instances upon which the incremental models are tested (e.g., in the Core50 dataset the last three sessions are designated as test fold \cite{lomonaco2017core50}).
\\
In principle, however, as each model $f(x|\theta_t)$ is estimated at time $t$ it should also be tested on data available at time $t$, possibly coming from a parallel (test) data stream. 
An idealised such scenario is one in which a person detector is continually trained on data coming from one surveillance camera, say in a shopping mall, and the resulting model is deployed (tested) in real time on all other cameras in the same mall. 

In our continual semi-supervised setting, model update cannot make use of any ground truth target values. 
If the latter are somehow available, but not shown to the learner for training purposes, the performance of the learner can in fact be evaluated on the stream of true target values by testing at each time $t$ model $f(x|\theta_t)$ on the data pair $(x_t,y_t)$.
A reasonable choice for a performance measure is the average loss of the series of models on the \emph{contemporary} data pair:
\begin{equation} \label{eq:performance}
\sum_{t = 1, \ldots, T} l( f(x_t|\theta_t), y_t ).
\end{equation}

In our experiments a standard 0/1 loss, which translates into classical accuracy, is used for classification tasks.

\section{Benchmark datasets} \label{sec:benchmarks}

To empirically validate CSSL approaches in a computer vision setting we created two benchmark datasets, one designed to test continual classification (CAR), and one for continual regression (CCC).

\subsection{Continual activity recognition (CAR) dataset}

\subsubsection{The MEVA dataset}

To allow the validation of continual semi-supervised learning approaches in a realistic classification task we created a new \emph{continual activity recognition} (CAR) dataset derived from the very recently released MEVA (Multiview Extended Video with Activities) dataset \cite{corona2021meva}. 

MEVA is part of the EctEV (Activities in Extended Video) challenge\footnote{\url{https://actev.nist.gov/}}. As of December 2019, 328 hours of ground-camera data and 4.2 hours of Unmanned Arial Vehicle video had been released, broken down into 4304 video clips, each 5 minutes long. These videos were captured at 19 sites (e.g. School, Bus station, Hospital) of the Muscatatuck Urban Training Center (MUTC), using a team of over 100 actors performing in various scenarios.  There are annotations for 22.1 hours (266 five-minute-long video clips) of data. The original annotations are available on GitLab\footnote{\url{https://gitlab.kitware.com/meva/meva-data-repo/tree/master/annotation/DIVA-phase-2/MEVA/}}.
Each video frame is annotated in terms of 37 different activity classes relevant to video surveillance (e.g. \textit{person\textunderscore opens\textunderscore facility\textunderscore door}, \textit{person\textunderscore reads\textunderscore document}, \textit{vehicle\textunderscore picks\textunderscore up\textunderscore person}). Each activity is annotated in terms of class labels and bounding boxes around the activity of interest. Whenever activities relate to objects or other persons (e.g., in \textit{person\textunderscore loads\textunderscore vehicle} the person usually puts an object into the vehicle’s trunk; in \textit{person\textunderscore talks\textunderscore to\textunderscore person} a number of people listen to the person speaking), these object(s) or people are also identified by a bounding box, to allow human-object interaction analyses.

\subsubsection{The CAR dataset}

The original MEVA dataset comes with a number of issues, from our standpoint: (i) multiple activity classes can take place simultaneously, whereas in our formulation (at least for now) only one label can be emitted at any given time instant; (ii) the quality of the original annotation is uneven, with entire instances of activities missing. As our tests are about classification, we can neglect the bounding box information.

For these reasons we set about creating our own continual activity recognition (CAR) dataset by selecting 45 video clips from MEVA and generating from scratch a modified set of annotations spanning a reduced set of 8 activity classes (e.g. \emph{person\_enters\_scene\_through\_structure}, \emph{person\_exits\_vehicle}, \emph{vehicle\_starts}). Those 8 classes have been suitably selected from the original 37 
to ensure that activity instances from different classes do not temporally overlap, so that we can assign a single label to each frame. Each instance of activity is annotated with the related start and end frame. 
Frames that do not contain any relevant activity label are assigned to a `background' class.
The goal is to classify the activity label of each input video frame.


For some MEVA sites, contiguous annotated videos exist with no gap between the end of the first video and the start of the second video. For some sites, ‘almost’ contiguous videos separated by short (5 or 15 min) gaps are available. Finally, videos from a same site separated by hours or days exist.
Accordingly, our CAR benchmark is composed of 15 sequences, broken down into three groups: 
\begin{enumerate}
\item
Five 15-minute-long sequences from sites G326, G331, G341, G420, and G638 formed by three original videos which are contiguous.
\item
Five 15-minute-long sequences from sites G329, G341, G420, G421, G638 formed by three videos separated by a short gap (5-20 minutes).
\item
Five 15-minute-long sequences from sites G420, G421, G424, G506, and G638 formed by three original videos separated by a long gap (hours or days).
\end{enumerate}
Each of these three evaluation settings is intended to simulate a different mix of continuous and discrete domain dynamics.

The CAR dataset including annotation and scripts is available on GitHub\footnote{\url{https://github.com/salmank255/IJCAI-2021-Continual-Activity-Recognition-Challenge}}.

\subsection{The continual crowd counting (CCC) dataset}

\emph{Crowd counting} is the problem of, given a video frame, counting the number of people present in the frame. While intrinsically a classification problem, crowd counting can be posed as a regression problem by manually providing for each training frame a density map \cite{boominathan2016crowdnet}. 
To date, crowd counting is mostly considered an image-based task, performed on single video frames. Few attempts have been made to extend the problem to the video domain \cite{hossain2020video}, \cite{fenget}, \cite{liuet}.

To the best of our knowledge, continual crowd counting has never been posed as a problem, not even in the fully supervised context -- thus, there are no standard benchmarks one can adopt in this domain.
For this reason we set about assembling the first benchmark dataset for \emph{continual crowd counting} (CCC).
Our CCC dataset is composed by 3 sequences, taken from existing crowd counting datasets:
\begin{enumerate}
\item
A single 2,000 frame sequence originally from the Mall dataset\footnote{\url{https://www.kaggle.com/c/counting-people-in-a-mall}} \cite{chen2012feature}.
\item
A single 2,000-frame sequence originally from the UCSD dataset\footnote{\url{http://www.svcl.ucsd.edu/projects/peoplecnt/}} \cite{chan2008privacy}.
\item
A 750-frame sequence from the Fudan-ShanghaiTech (FDST) dataset\footnote{\url{https://drive.google.com/drive/folders/19c2X529VTNjl3YL1EYweBg60G70G2D-w}}, 
composed by 5 clips, 150 frames long, portraying a same scene \cite{fang2019locality}.
\end{enumerate}
The ground truth for the CCC sequences (in the form of a density map for each frame) was generated by us for all three datasets following a standard annotation protocol\footnote{\url{https://github.com/svishwa/crowdcount-mcnn}}.

The CCC dataset complete with scripts for download and generating the baseline results is available on GitHub\footnote{\url{https://github.com/Ajmal70/IJCAI_2021_Continual_Crowd_Counting_Challenge}}.

\section{Challenges} \label{sec:challenges}

\subsection{Protocols}

\subsubsection{CAR challenge}

From our problem definition (Sec. \ref{sec:continual-ssl}), once a model is fine-tuned on the supervised portion of a data stream it is then both incrementally updated using the unlabelled portion of the same data stream and tested there, using the provided ground truth.
Incremental training and testing are supposed to happen independently for each sequence, as the aim is to simulate real-world scenarios in which a smart device with continual learning capability can only learn from its own data stream. 
However, as CAR sequences do not contain instances of all 9 activities, the initial supervised training is run there on the union of the supervised folds for each sequence.

\emph{Split}. 
Each data stream (sequence) in our benchmark(s) is divided into a supervised fold ($S$), a validation fold ($V$) and a test fold ($T$), the last two unsupervised.
Specifically, in CAR given a sequence the video frames from the first five minutes (5 x 60 x 25 = 7,500 samples) are selected to form the initial supervised training set $S$.
The second clip (another 5 minutes) is provided as validation fold to tune the 
CSSL strategy.
The remaining clip (5 minutes) is used as test fold for testing the performance of the (incrementally updated) classifier. 

\emph{Evaluation}. 
Performance is evaluated as the average performance of the incrementally updated classifier over the test fold for all the 15 sequences.
More specifically, we evaluate the average 
accuracy (percentage of correctly classified frames) over the test folds of the 15 sequences.
Remember that, however, in our setting each test frame
is classified by the \emph{current} model available at time $t$ \emph{for that specific sequence}.
This distinguishes our evaluation setting from classical ones in which all samples at test time are processed by the same model, whose accuracy is then assessed.

\subsubsection{CCC challenge}
Unlike the CAR challenge, in the crowd counting case (as this is a regression problem) each sequence is treated completely independently.

\emph{Split}.
For the CCC challenge we distinguish two cases. For the 2,000-frame sequences from either the UCSD or the Mall dataset, $S$ is formed by the first 400 images, $V$ by the following 800 images, and $T$ by the remaining 800 images. For the 750-frame sequence from the FDST dataset, $S$ is the set of the first 150 images, $V$ the set of the following 300 images, and $T$ the set of remaining 300 images.

\emph{Evaluation}.
For continual crowd counting, MAE (Mean Absolute Error) is adopted (as standard in the field) to measure performance. MAE is calculated using predicted and ground truth density maps in a regression setting. 

\begin{figure*}[h]
    \centering
    \includegraphics[width=1\textwidth]{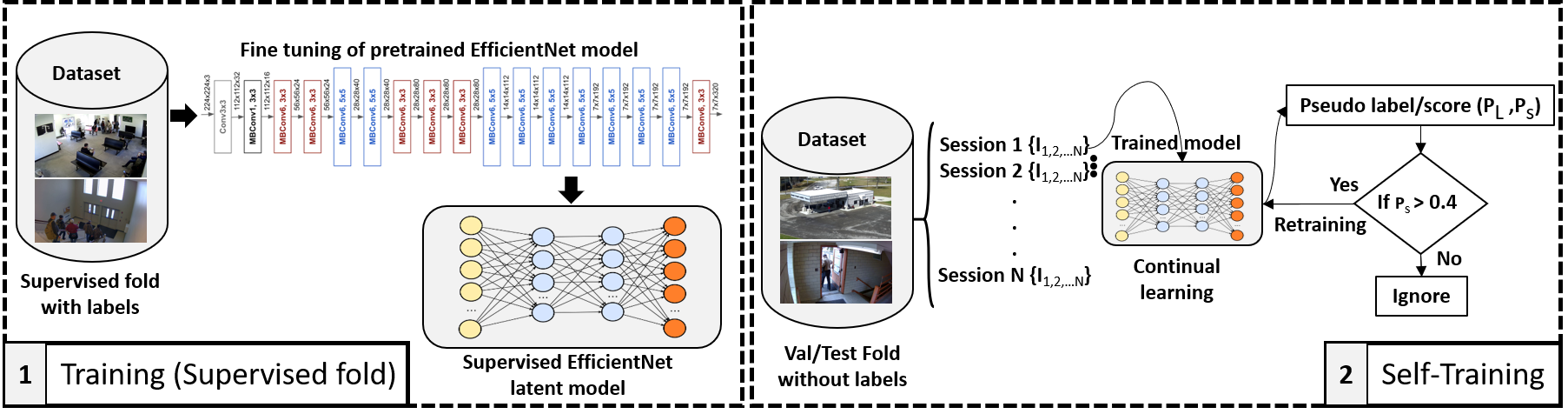}
    \caption{{Overall pipeline of our CAR baseline. 1) Firstly, a pre-trained EfficientNet model is fine-tuned over the supervised fold of the dataset (jointly over the 15 sequences). 
    2) The unlabeled validation and test folds are divided into subfolds. In each session 
    video frames from each sub-fold are used in a self-training cycle in which a pseudo label $P_L$ with prediction probability score $P_s$ is generated for each frame, but only the pseudolabels with 
    $P_s$ above a fixed threshold (namely, 0.4) are considered as ground truth for retraining the model. Frames with 
    lower $P_s$ are ignored. 
    The model updated by self-training in session $n$ is used as initial model in session $n+1$.
    }}
    \label{fig:CAR}
\end{figure*}

\begin{table*}
\begin{center}
\footnotesize \scalebox{1} {
\begin{tabular}{|*{3}{p{0.036\textwidth}}|*{6}{p{0.036\textwidth}}|*{6}{p{0.036\textwidth}}|}
\hline
 \multicolumn{3}{|c}{}& \multicolumn{6}{c|}{\textbf{Validation Fold}}  & \multicolumn{6}{c|}{\textbf{Test Fold}}\\
\hline
\multicolumn{3}{|c|}{Methods / Evaluation metrics}  & \multicolumn{2}{c|}{Precision} & \multicolumn{2}{|c|}{Recall} & \multicolumn{2}{c|}{F1-Score} & \multicolumn{2}{c|}{Precision} & \multicolumn{2}{|c|}{Recall} & \multicolumn{2}{c|}{F1-Score} \\
\multicolumn{3}{|c|}{}  & C & \multicolumn{1}{c|}{W} & C & \multicolumn{1}{c|}{W} & C & \multicolumn{1}{c|}{W} & C & \multicolumn{1}{c|}{W} & C & \multicolumn{1}{c|}{W} & C & W \\
\hline
\multicolumn{3}{|l|}{$sup-ft-union$}  & \textbf{0.20} & \textbf{0.74} & \textbf{0.25} & 0.68 & \textbf{0.16} & 0.70 & \textbf{0.16} & \textbf{0.82} & \textbf{0.15} & 0.77 & \textbf{0.14} & 0.79  \\
\multicolumn{3}{|l|}{$upd-V$ / $upd-T$}  & 0.18 & 0.73 & 0.19 & \textbf{0.73} & 0.14 & \textbf{0.72} & 0.14 & 0.81 & \textbf{0.15} & \textbf{0.81} & \textbf{0.14} & \textbf{0.80}  \\
\multicolumn{3}{|l|}{$upd-V+T$}  & 0.17 & 0.73 & 0.19 & \textbf{0.73} & 0.14 & \textbf{0.72} & 0.14 & 0.80 & 0.14 & 0.79 & 0.13 & 0.79  \\

\hline
\end{tabular}
}
\end{center}

\caption{Performance of 
the initial supervised model versus that of the incrementally updated models (separately on $V$ and $T$ or in combination), using
three standard evaluation metrics. For each metric we report both class average (C) and weighted average (W).}
\label{tab:tab1}
\end{table*}

\subsection{Tasks}

For our IJCAI challenge we therefore set four different validation experiments (\emph{tasks}).
\begin{enumerate}
\item
In the first task (\emph{CAR-absolute}) the goal is to achieve the best average performance across all the (test folds of the) 15 sequences in the CAR dataset. The choice of the baseline action recognition model is left to the participants.
\item
In the second task (\emph{CAR-incremental}) the goal is to achieve the best performance differential between the model as updated through the chosen CSSL strategy and the original model fine-tuned over the supervised fold. 
We thus evaluate
the difference between the average performance of the incrementally updated model on the test fold and the average performance of initial model, 
also on the test fold. The baseline recognition model is set by us (see Baselines).
\item
Task \emph{CCC-absolute} seeks the best average performance over the 
test fold of the 3 sequences of the CCC dataset. The choice of the baseline crowd counting model is left to the participants to encourage them to push its performance to the limit.
\item
Finally, task \emph{CCC-incremental} seeks the best 
performance differential between the initial and the updated model over the test fold, averaged across the three sequences.
The baseline crowd counting model is set.
\end{enumerate}

\subsection{Baselines}

To provide a baseline for the above tasks, and assess the degree of difficulty of the challenges,
we decided to adopt a simple strategy which consists of classical (batch) semi-supervised self-training \emph{performed in a series of sessions}, for both the activity and the crowd counting experiments.
Baselines are provided regarding the initial action recognition model to be used in the CAR tests, the base crowd counter to be used in the CCC tests, and the semi-supervised incremental learning process itself.

\subsubsection{Baseline activity recognition model}

The baseline model is the recent EfficientNet network 
(model EfficientNet-B5) \cite{tan2019efficientnet}, pre-trained on the large-scale ImageNet dataset over 1000 classes. For our tests we initialised the model using those weights and changed the number of classes to 9 activities (see Section \ref{sec:benchmarks}). Detailed information about its implementation, along with pre-trained models, can be found on Github\footnote{\url{https://github.com/lukemelas/EfficientNet-PyTorch}}. The model is easily downloadable using the Python command “pip” (\textit{pip install efficientnet-pytorch}).

\begin{table*}[h!]
\centering
\footnotesize
\begin{tabular}{ |p{2.5cm}|p{1.0cm}|p{1.8cm}|p{1.8cm}|p{1.8cm}|p{1.8cm}|p{1.8cm}|  }
 \hline
 \multicolumn{4}{|c|}{\textbf{Validation Fold $V$}} & 
 \multicolumn{3}{|c|}{\textbf{Test Fold $T$}}  \\ 
 \hline
 
 & $FDST$ & $UCSD$ & $MALL$  & $FDST$ & $UCSD$ & $MALL$  \\
\hline
$sup-ft$  & \textbf{5.17} & \textbf{6.36}   &  \textbf{6.65}    & 9.59     & 8.34  & 13.56     \\ [0.5ex]

$upd-V$ &   6.05 &  7.58   & 7.40 & \textbf{8.00} & \textbf{7.45} & 16.38   
\\ [0.5ex] 

$upd-V+T$   &   
8.32 &   
8.28
& 
9.36
& 
9.16
&  
8.93
& 
\textbf{9.46}
\\ [0.5ex] 

 \hline
\end{tabular}
\caption{Evaluation on the CCC validation ($V$) and test ($T$) folds for different experimental setups. Here $sup-ft$, $upd-V$, and $upd-V+T$ refer to the supervised model fine-tuned on $S$, the model incrementally updated via self-training in sessions on the validation fold $V$, and the model incrementally updated on the combined validation $V$ and test $T$ folds, respectively. The average test MAE is reported for each model.}
\label{table:val}
\end{table*}

\subsubsection{Baseline crowd counter}

For the baseline crowd counting model, we selected the Multi-Column Convolutional Neural Network (MCNN) \cite{zhang2016single}, whose (unofficial) implementation is publicly available\footnote{\url{https://github.com/svishwa/crowdcount-mcnn}} and uses PyTorch. {MCNN was considered state-of-the-art when released in 2016, but is still commonly used as a standard baseline in more recent works, due to its competitive performance on public datasets (e.g., \cite{Jinag2019,Xiong2019}. MCNN made significant contributions to crowd counting by proposing a network architecture better-equipped to deal with differing head sizes due to image resolution, perspective effects, distances between the camera and the people within the scene, etc.  In a nutshell, this robustness to scale was achieved via a network composed of three parallel CNN columns, each of which used filters with different receptive field sizes, allowing each column to specialize for a particular scale of human head.}

Pre-trained MCNN models are available for both the ShanghaiTech A and the ShanghaiTech B datasets. For our tests, as well as the Challenge, we chose to adopt the ShanghaiTechB pre-trained model.

\subsubsection{Baseline incremental learning strategy}

As mentioned, our baseline for incremental learning from unlabelled data stream is instead based on a \emph{vanilla self-training} approach \cite{Trigueroet}.
For each sequence, the unlabelled data stream (without distinction between validation and test folds) is partitioned into a number of sub-folds. Each sub-fold spans 1 minute in the CAR challenges, so that each unlabelled sequence is split into 10 sub-folds. Sub-folds span 100 frames in the CCC challenges, so that the UCSD and MALL sequences comprise 16 sub-folds whereas the FDST sequence contains only 6 sub-folds.

Starting with the model initially fine-tuned on the supervised portion of the data stream, self-training is successively applied in a batch fashion in sessions, one session for each sub-fold, for a single epoch (as we noticed that using multiple epochs would lead to results degradation). {Self-training requires to select a confidence threshold above which predictions are selected as pseudolabels.
In our tests we 
set a confidence threshold of 0.4.} The predictions generated by the model obtained after one round of self-training
upon a sub-fold are stored as baseline predictions for the current sub-fold. The model updated after each self-training session is used as initial model for the following session.

\section{Results} \label{sec:experiments}

\subsection{Results on Activity Recognition}

Three types of experiments were performed to measure: (i) the performance of the baseline model after fine-tuning on the union of the supervised folds for the 15 CAR sequences ($sup-ft-union$), without any continual learning; (ii) the performance obtained by incrementally updating the model (in sessions) using the unlabelled validation ($upd-V$) or test ($upd-T$) data streams, considered separately; (iii) the performance of self-training over validation and training folds considered as a single contiguous data stream, again in a session-wise manner ($upd-V+T$).

The results are shown in Table \ref{tab:tab1}.
The standard classification evaluation metrics precision, recall, and F1-score were used for evaluation. For each evaluation metric we computed both the \emph{class average} (obtained by computing the score for each class and taking the average 
without considering the number of training samples in each class, i.e., $F1_{class1} + F1_{class2} \cdots + F1_{class9} $ ) and the \emph{weighted average} (which uses the number of samples $W_i$ in each class $i$, i.e., $F1_{class1}*W_1 + F1_{class2}*W_2 \cdots + F1_{class9}*W_9 $ ). 

It can be noted that on the validation fold continual updating does improve performance to some extent, especially under Recall and F1 score. Improvements in Recall are visible on the test fold as well. All in all the baseline appears able to extract information from the unlabelled data stream.

\subsection{Results on Crowd Counting}

Table \ref{table:val} shows a quantitative analysis of the performance of the fine-tuned supervised model ($sup$) and two incrementally updated models ($upd$) on the validation and the test split, respectively, using the mean absolute error (MAE) metric. 

The experiments were performed under three different experimental settings: (i) the initial model fine tuned on the supervised folds ($sup-ft$), (ii) a model incrementally updated on the validation fold ($upd-V$), and (iii) a model incrementally updated on both the validation and test folds considered as a single data stream ($upd-V+T$). In Table \ref{table:val} all three models are tested on both the $V$ and $T$ portions of the data streams.
The initial model was trained for 100 epochs on the supervised fold ($S$), whereas the incremental models were self-trained for 5 epochs in each session (we also tried 2 epochs with no significant change observed). 


{When compared with classical papers in the crowd counting literature, the MAE values in Table \ref{table:val} are noticeable higher.  Concretely, a recent work that incorporates optical flow to improve crowd counting achieved MAEs of 1.76, 0.97, and 1.78 on the FDST, UCSD, and Mall dataset, respectively \cite{hossain2020video}.  Also, the original MCNN implementation yielded MAEs of 3.77 and 1.07 on FDST and UCSD (performance was not reported on Mall) \cite{zhang2016single,fang2019locality}. 
The higher MAEs reported in our work are expected,  due to the significantly different and more challenging training protocol (i.e., batch training in standard crowd counting work vs.\ continual learning here).  For example, standard papers in crowd counting following the typical evaluation protocol employ 800 training images for the Mall and UCSD datasets; whereas, in our problem setting 
only 400 images are used for supervised training. }

Previous batch results should be seen as upper bounds to our incremental task, whereas ours should be considered as new baselines for a new problem setting.
While model updating does not seem to be effective (on average) on the validation streams, the effect is quite significant on the test fold (right side of the Table), with important performance gains.



\section{Conclusions} \label{sec:conclusions}

In this paper we formulated the continual semi-supervised learning problem and proposed suitable training and testing protocols for it. To encourage future work in the area we created the first two benchmark datasets, as the foundations of our IJCAI 2021 challenge. We proposed a simple strategy based on batch self-training a baseline model in sessions. The results show that, in both the activity recognition and the crowd counting challenge, the baseline appears in fact to be able to extract information from the unlabelled data stream. 
Nevertheless, the need for a more sophisticated approach leveraging the dynamics of the data stream is clear.

\bibliographystyle{named}
\bibliography{ijcai21}

\end{document}